\documentclass{article}

\usepackage{microtype}

\usepackage{graphicx}
\usepackage{subfigure}
\usepackage{booktabs} %
\usepackage{xspace}

\usepackage[dvipsnames,table]{xcolor}        %
\colorlet{highlight}{BurntOrange!15}

\usepackage[normalem]{ulem}

\usepackage[frozencache]{minted}
\definecolor{LightGray}{gray}{0.9}

\usepackage{hyperref}

\usepackage{siunitx}
\robustify\textbf
\robustify\bfseries
\robustify\uline

\usepackage[accepted]{icml2024}

\usepackage{subcaption}
\usepackage{amsmath}
\usepackage{amssymb}
\usepackage{mathtools}
\usepackage{amsthm}

\usepackage[capitalize,noabbrev]{cleveref}

\theoremstyle{plain}

\theoremstyle{definition}

\theoremstyle{remark}

\usepackage{tikz}

\usetikzlibrary{external}
\usetikzlibrary{
  calc,
  fit,
  matrix,
  patterns,
  patterns.meta,
  positioning
}

\usepackage{pgfplots}
\usepgfplotslibrary{
  colorbrewer,
}

\makeatletter
\patchcmd{\NAT@test}{\else \NAT@nm}{\else \NAT@nmfmt{\NAT@nm}}{}{}

\DeclareRobustCommand\citepos
  {\begingroup
   \let\NAT@nmfmt\NAT@posfmt%
   \NAT@swafalse\let\NAT@ctype\z@\NAT@partrue
   \@ifstar{\NAT@fulltrue\NAT@citetp}{\NAT@fullfalse\NAT@citetp}}

\let\NAT@orig@nmfmt\NAT@nmfmt
\def\NAT@posfmt#1{\NAT@orig@nmfmt{#1's}}
\makeatother

\usepackage{enumitem}

\usepackage{xspace}
\makeatletter
\DeclareRobustCommand\onedot{\futurelet\@let@token\@onedot}
\def\@onedot{\ifx\@let@token.\else.\null\fi\xspace}

\def\ie{{i.e}\onedot} 
\def\cf{{cf}\onedot} \def\Cf{{Cf}\onedot}
 
\def\wrt{w.r.t\onedot}

\makeatother

\hypersetup{
  colorlinks,
  linkcolor = BrickRed,
  citecolor = RoyalBlue,
  urlcolor  = WildStrawberry,
}

\makeatletter
\newcommand{\methodname}{MaSSL\@ifnextchar.{\@}{\xspace}}
\makeatother

\icmltitlerunning{Learning from Memory: Non-Parametric Memory Augmented Self-Supervised Learning of Visual Features}

\begin{document}

\twocolumn[
\icmltitle{Learning from Memory: Non-Parametric Memory Augmented Self-Supervised Learning of Visual Features}

\begin{icmlauthorlist}
\icmlauthor{Thalles Silva}{aaa}
\icmlauthor{Helio Pedrini}{aaa}
\icmlauthor{Ad\'in Ram\'irez Rivera}{bbb}
\end{icmlauthorlist}

\icmlaffiliation{aaa}{
Institute of Computing, University of Campinas, Campinas-SP, Brazil}
\icmlaffiliation{bbb}{
Department of Informatics, University of Oslo, Oslo, Norway}

\icmlcorrespondingauthor{Thalles Silva}{thalles.silva@students.ic.unicamp.br}
\icmlcorrespondingauthor{Helio Pedrini}{helio@ic.unicamp.br}
\icmlcorrespondingauthor{Ad\'in Ram\'irez Rivera}{adinr@uio.no}

\icmlkeywords{Machine Learning, ICML}

\vskip 0.3in
]

\printAffiliationsAndNotice{}  %

\begin{abstract}
This paper introduces a novel approach to improving the training stability of self-supervised learning (SSL) methods by leveraging a non-parametric memory of seen concepts. The proposed method involves augmenting a neural network with a memory component to stochastically compare current image views with previously encountered concepts. Additionally, we introduce stochastic memory blocks to regularize training and enforce consistency between image views. We extensively benchmark our method on many vision tasks, such as linear probing, transfer learning, low-shot classification, and image retrieval on many datasets. The experimental results consolidate the effectiveness of the proposed approach in achieving stable SSL training without additional regularizers while learning highly transferable representations and requiring less computing time and resources. Code at \url{https://github.com/sthalles/MaSSL}.
\end{abstract}

\section{Introduction}

Self-supervised learning (SSL) based on join-embedding architectures currently holds state-of-the-art performance on many representation learning benchmarks.
Among different methods, clustering-based approaches \citep{caron2021emerging,caron2018deep,caron2019unsupervised,asano2019self,Silva2023} appear to be the most successful recipe for learning self-supervised features in the visual domain.
SSL clustering methods learn image representations by discretizing the embedding space.
They set up optimization tasks that involve learning a finite set of prototypes or centroids based on the self-supervised signal coming from views of an image.
Despite their high ability to learn representations, clustering methods are notoriously difficult to train due to their susceptibility to training collapse.

We argue that learning the prototypes via gradient descent is the primary source of poor training instability in SSL clustering methods.
Due to the lack of human labels and excessive noise from the self-supervised signals, the network attempts to cluster all the embeddings into a single prototype as the most efficient way to optimize the loss function.
Current methods avoid collapse by employing additional regularizers that force representations to spread evenly in the space over the prototypes.
Examples include: (1)~the combination of centering and target sharpening \citep{caron2021emerging, zhou2021ibot}, (2)~the mean entropy maximization (ME-MAX) \citep{assran2021semisupervised, Silva2023}, and (3)~the Sinkhorn-Knopp \citep{asano2019self,caron2020unsupervised}.
In addition, state-of-the-art SSL methods based on Vision Transformers (ViTs) use the full output of the Transformer (\texttt{[CLS]} + patch token) and optimize the MIM (Masked Image Modeling) pretext task on the patch embeddings \citep{zhou2021ibot,oquab2023dinov2}.
Despite performance gains, this architectural choice drastically increases computational costs and training time.

Motivated by the current landscape of SSL methods, we propose a new stable method that exceeds current approaches on many retrieval and transfer tasks while reducing computing resources and training time.
Based on the intuition that learning relies on memory, we present a non-parametric approach that poses the SSL problem in terms of learning from past experiences.
We augment a neural network with a memory component that holds a snapshot of the most recent image representations seen by the model.
Unlike previous approaches that use memory/queues to mine negatives \citep{he2020momentum} or positives \citep{Dwibedi_2021_ICCV} in a contrastive learning setup, our proposed memory allows the network to learn visual representations by comparing current events (views of an image) with previously experienced concepts (image representations from previous iterations) in memory.
We named this method \textbf{M}emory \textbf{A}ugmented \textbf{S}elf-\textbf{S}upervised \textbf{L}earning (MaSSL).

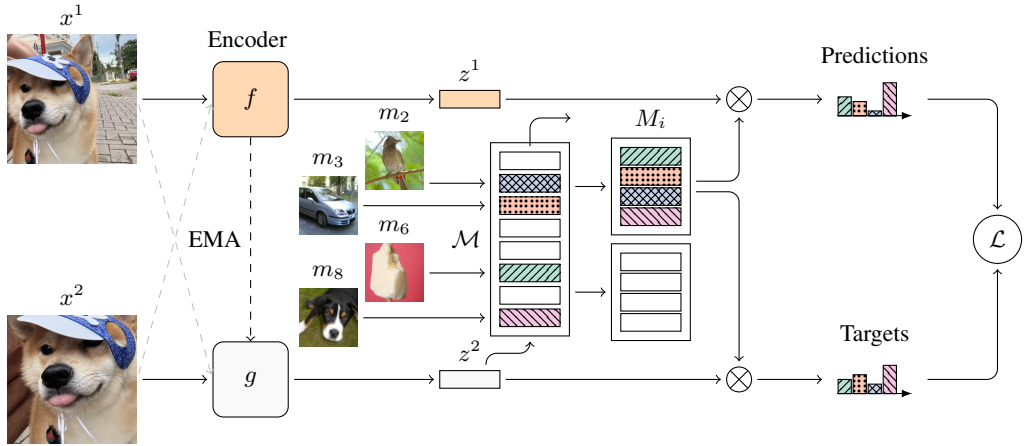
\begin{figure*}
    \centering
    \begin{tikzpicture}[
      module/.style={
        rectangle,
        rounded corners,
        minimum width=1.0cm,
        minimum height=1cm,
        text centered,
        draw=black,
      },
      representation/.style={
        rectangle,
        minimum width=0.8cm,
        minimum height=0.1cm,
        draw=black,
      },
      arrow/.style={
        ->,
        shorten >=2pt,
        shorten <=2pt,
        rounded corners,
      },
      font=\footnotesize,
      cross/.style={path picture={
          \draw[black]
        (path picture bounding box.south east) -- (path picture bounding box.north west) (path picture bounding box.south west) -- (path picture bounding box.north east);
      }}
    ]

      \node[inner sep=0pt, label=above:$x^1$] (view1) {\includegraphics[width=.1\textwidth]{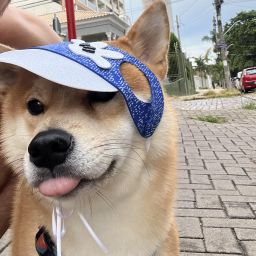}};
      \coordinate[below=of view1] (midpoint);

      \node[inner sep=0pt, below=of midpoint, label=above:$x^2$] (view2) {\includegraphics[width=.1\textwidth]{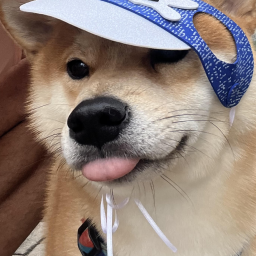}};

      \node [module, fill=orange!30, right=of view1, label=above:Encoder\phantom{j}] (enc1) {$f$};
      \node [module, fill=gray!5, right=of view2] (enc2) {$g$};

      \coordinate[right=of enc1] (proj1);
      \coordinate[right=of enc2] (proj2);

      \node [representation, draw, fill=orange!30, right=of proj1, label=above:{$z^1$}] (z1)  {};

      \node [representation, draw, fill=gray!5, right=of proj2, label=above:{$z^2$}] (z2)  {};

      \coordinate[right=of z1] (assigner1);
      \coordinate[right=of z2] (assigner2);

      \matrix [draw, row sep=0.5mm, xshift=8mm, label=left:{$\mathcal{M}$}] (memory) at (midpoint-|z1)
      {
        \node[representation](m1) {}; \\
        \node[representation, preaction={fill, Set2-C!50}, pattern={crosshatch}](m2){}; \\
        \node[representation, preaction={fill, Set2-B!50}, pattern={Dots[distance=2pt]}](m3){}; \\
        \node[representation](m4) {}; \\
        \node[representation](m5) {}; \\
        \node[representation, preaction={fill, Set2-A!50}, pattern={north east lines}](m6) {}; \\
        \node[representation](m7) {}; \\
        \node[representation, preaction={fill, Set2-D!50}, pattern={north west lines}] (m8) {}; \\
      };

      \coordinate[below=0.25 of assigner1] (dequeue);
      \draw[arrow] (m1.north) |- (dequeue.west);

      \node[inner sep=0pt, label=above:$m_2$, left=of m1] (im_m1) {\includegraphics[width=.045\textwidth]{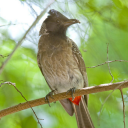}};

      \node[inner sep=0pt, xshift=-25pt, label=above:$m_3$, left=of m3] (im_m4) {\includegraphics[width=.045\textwidth]{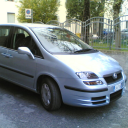}};

      \node[inner sep=0pt, label=above:$m_6$, left=of m6] (im_m6) {\includegraphics[width=.045\textwidth]{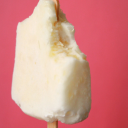}};

      \node[inner sep=0pt, xshift=-25pt, label=above:$m_8$, left=of m8] (im_m8) {\includegraphics[width=.045\textwidth]{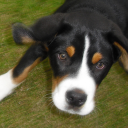}};

      \draw[arrow] (im_m1.south east |- m2.west) -- (m2.west -| memory.west);
      \draw[arrow] (im_m4.east) -- (m3.west -| memory.west);
      \draw[arrow] (im_m6.east) -- (m6.west -| memory.west);
      \draw[arrow] (im_m8.east) -- (m8.west -| memory.west);

      \coordinate[right=of assigner1] (dist1);
      \coordinate[right=of assigner2] (dist2);

      \matrix [draw, row sep=0.2mm, yshift=20, label=above:{$M_i$}] (block1) at (midpoint-|dist1)
      {
        \node[representation, preaction={fill, Set2-A!50}, pattern={north east lines}] {}; \\
        \node[representation, preaction={fill, Set2-B!50}, pattern={Dots[distance=2pt]}] {}; \\
        \node[representation, preaction={fill, Set2-C!50}, pattern={crosshatch}] {}; \\
        \node[representation, preaction={fill, Set2-D!50}, pattern={north west lines}] {}; \\
      };

      \matrix [draw, row sep=0.2mm, yshift=-20] (block2) at (midpoint-|dist1)
      {
        \node[representation] {}; \\
        \node[representation] {}; \\
        \node[representation] {}; \\
        \node[representation] {}; \\
      };

      \node [circle, draw, cross, right=of dist1] (t1)  {};
      \node [circle, draw, cross, right=of dist2] (t2)  {};

      \node[right=of t1, label=above:Predictions](pred_dist){
          \begin{tikzpicture}[ybar, bar width=5pt,
          ]
              \draw[preaction={fill, Set2-A!50}, pattern={north east lines}]
                  plot coordinates{(0,4/16)(0.2,0)(0.4,0)(0.6,0)};
              \draw[preaction={fill, Set2-B!50}, pattern={Dots[distance=2pt]}]
                  plot coordinates{(0,0)(0.2,3/16)(0.4,0)(0.6,0)};
              \draw[preaction={fill, Set2-C!50}, pattern={crosshatch}]
                  plot coordinates{(0,0)(0.2,0)(0.4,1/16)(0.6,0)};
              \draw[preaction={fill, Set2-D!50}, pattern={north west lines}]
                  plot coordinates{(0,0)(0.2,0)(0.4,0)(0.6,7/16)};

              \draw (-0.1,0) edge[-latex] (0.9,0);
          \end{tikzpicture}
      };

      \node[right=of t2, label=above:Targets](target_dist){
          \begin{tikzpicture}[ybar, bar width=5pt,
          ]
              \draw[preaction={fill, Set2-A!50}, pattern={north east lines}]
              plot coordinates{(0,3/16)(0.2,0)(0.4,0)(0.6,0)};
              \draw[preaction={fill, Set2-B!50}, pattern={Dots[distance=2pt]}]
              plot coordinates{(0,0)(0.2,4/16)(0.4,0)(0.6,0)};
              \draw[preaction={fill, Set2-C!50}, pattern={crosshatch}]
              plot coordinates{(0,0)(0.2,0)(0.4,2/16)(0.6,0)};
              \draw[preaction={fill, Set2-D!50}, pattern={north west lines}]
              plot coordinates{(0,0)(0.2,0)(0.4,0)(0.6,6/16)};

              \draw (-0.1,0) edge[-latex] (0.9,0);
          \end{tikzpicture}
      };

      \draw[arrow] (view1.east) -- (enc1.west);
      \draw[arrow] (view2.east) -- (enc2.west);

      \draw[arrow, dashed, black!25] (view1.east) -- (enc2.west);
      \draw[arrow, dashed, black!25] (view2.east) -- (enc1.west);

      \draw[arrow] (enc1.east) -- (z1.west);
      \draw[arrow] (enc2.east) -- (z2.west);

      \draw[arrow] (z1.east) -- (t1.west);
      \draw[arrow] (z2.east) -- (t2.west);

      \draw[arrow] (t1.east) -- (pred_dist.west);
      \draw[arrow] (t2.east) -- (target_dist.west);

      \draw[arrow] ([yshift=20pt] memory.east) -- (block1.west);
      \draw[arrow] ([yshift=-20pt] memory.east) -- (block2.west);

      \draw[arrow] ([yshift=2pt] block1.east) -| (t1.south);
      \draw[arrow] ([yshift=-2pt] block1.east) -| (t2.north);

      \draw[arrow] (z2.30) |- ($(z2.30)!.5!(memory.south)$) -| (memory.south);

      \path[->, dashed] (enc1) edge node [midway, left] {EMA} (enc2);

      \coordinate[right=of pred_dist] (pred-right);
      \node [circle, draw,] (loss) at (memory -| pred-right)  {$\mathcal{L}$};
      \draw[arrow] (pred_dist.east) -| (loss.north);
      \draw[arrow] (target_dist.east) -| (loss.south);

    \end{tikzpicture}
    \caption{Learning from memory. Given two or more views of an image, each view is encoded by the student and teacher encoders, resulting in respective vector representations $z^1$ and $z^2$. Each view's representation is compared against representations of previously seen images in memory, resulting in respective similarity distributions. Note that the working memory $\mathcal{M}$ is split into blocks, $M_i$, of randomly chosen representations. The learning objective, $\mathcal{L}$, forces the similarity distributions of views \wrt the memory representations to be consistent. In a case where the model perceives an image of a dog, the interaction between what it currently sees and what it remembers should produce (1)~strong similarity scores for previously seen dogs, (2)~weak scores for non-related images in the memory, and (3)~interactions should be consistent among views.}
    \label{fig:overview}
\end{figure*}

In addition to the working memory, we introduce the concept of stochastic memory blocks.
Stochastic blocks allow the network to retrieve a random subset of representations from previous iterations.
These representations symbolize concepts previously seen by the model and are used as anchors to enforce consistency between the current image views.
We show that stochastic memory blocks regularize the learning problem, making our method stable even without additional regularizers to prevent mode collapse.
Finally, our loss optimizes for consistency between views of an image by matching their view-memory similarity distributions, which means that views of an image must activate similar memory representations with similar scores.
In other words, views should output consistent similarity patterns when compared to representations of other images in the memory. \Cref{fig:overview} presents a pictorial overview of our learning architecture.

Our contributions are threefold:
\begin{itemize}[nosep, leftmargin=*]
    \item A novel SSL pretext task that learns visual representations by formulating multiple discriminative tasks based on comparing the current perceived signal to previously experienced concepts stored in memory.
    \item A stochastic memory, implemented through a non-parametric distribution of the past image representations and a memory block mechanism that allows representation learning in a prototype-free manner.
    \item A simple SSL learning framework that does not require additional regularizers to avoid training collapse and operates on the \texttt{[CLS]} token of the ViT, reducing the pre-training time and memory requirements while learning highly transferable representations.
\end{itemize}

\section{Related Work}

\textbf{Self-supervised learning} has evolved from more specialized pretext tasks such as solving rotations \citep{gidaris2018unsupervised}, jigsaw puzzles \citep{noroozi2016unsupervised}, and relative positions \citep{doersch2015unsupervised}, to a predominant set of tasks based on instance discrimination \citep{he2020momentum,chen2020simple,chen2021exploring,silva2023clove}. Current methods mainly differ from one another on (1)~how they avoid mode collapse and (2)~how they pose the view-invariance task, which may be embedding- \citep{grill2020bootstrap} or clustering-based \citep{caron2020unsupervised}. Current state-of-the-art SSL is based on the Transformer architecture \citep{dosovitskiy2020image}. Some approaches formulate their loss function over the \texttt{[CLS]} token only \citep{caron2021emerging}, while the most recent and powerful methods use the full output of the Transformer, \ie, \texttt{[CLS]} + patch tokens \citep{zhou2021ibot,oquab2023dinov2}.

\textbf{Memory banks or queues} in SSL are not new concepts.
Many proposed techniques \citep{misra2020self,chen2021mocov3} rely on storing representations in a container, often called support set or queues.
In MoCo \citep{chen2021mocov3} and PIRL \citep{misra2020self}, the memory is used as a source of negative representations, \ie, the currently processed image is pushed away from distinct image representations in a queue in a contrastive task by minimizing the InfoNCE \citep{oord2018representation} loss.
Alternatively, \citet{Dwibedi_2021_ICCV} uses an extra queue as a source of positives.
Specifically, the support set is used to bootstrap nearest neighbors for the current views, framing a contrastive learning task that approximates views of an image to their neighbors' representations.

\methodname uses the memory container differently.
To begin, \methodname is a negative-free contrastive\footnote{The literature uses the misnomer ``non-contrastive'' to refer to methods that do not explicitly use negative examples while learning the representations. We, however, argue that there is a better way of naming these methods.} method.
Hence, it does not need to bootstrap negatives for its learning objective, nor does it have an explicit term in the loss function to push representations apart and avoid collapse.
Most importantly, \methodname uses the memory to formulate discriminative tasks.
Intuitively, if the currently perceived image is similar to one or more images the model has seen (in memory), they should relate with a strong similarity score.
Conversely, if the current image is not semantically similar to one or more images in the memory, they should relate with a weak similarity score.
On top of that, \methodname's learning objective forces multiple views of the same image to agree on how they relate to previously perceived representations.

Similar to \methodname, \citet{assran2021semisupervised} proposed a semi-supervised method, termed PAWS, that employs a support set composed of uniformly distributed labeled examples as anchors to optimize for views' consistency.
While \methodname may be regarded as an SSL version of PAWS, translating the learning problem in the former to an SSL setup is not trivial.
PAWS takes advantage of the additional, \textit{free-of-noise} signal to incorporate biases into the learning problem, stabilizing the training process.
In addition to human labels, PAWS uses a regularizer to spread views' assignments over the examples in the support set.
In contrast, \methodname does not require human-labeled examples to learn visual representations, nor does it employ regularizers to prevent collapse.

\section{Methodology}

Inspired by how humans recall and generalize observations based on memory comparisons, we introduce a memory (based on a non-parametric distribution) to our methods, allowing the network to contrast representations from current events to previous iterations.
In addition, we regularize the learning process using a stochastic memory partition strategy, forcing the representations to be general and not susceptible to particular shortcuts.
\Cref{fig:overview} depicts an overview of our proposed method.

\textbf{Notation.} Let $x$ be an image from a large unlabeled dataset $\mathcal{X}$ and $x^v$ a view of $x$, for $v=1,2, \dots, N_v$.
We define $f$ and $g$ as a pair of student and teacher vision encoders where the student is trained with backpropagation, and the teacher is distilled through exponential moving average (EMA) from the student.
Each encoder receives a view, $x^v$, and produces a low-dimensional representation $z^v$, such that $z^v_s=f(x^v)$ and $z^v_t=g(x^v)$, where the subscripts $s$ and $t$ represent student and teacher branches.

In addition, we denote by $\mathcal{M} = \left \{ m_k \right \}_{k=0}^K$ a vector container designed to simulate a working memory that temporarily stores vector representations from images previously seen during training by the model.

\subsection{Memory}
\label{sec:memory}
When humans experience something for the first time, there is likely an additional excitement or surprise due to encountering novelty.
When a similar experience happens again, however, the surprise will probably not be the same.
This occurs because of memory and its essential role in learning.
Indeed, the fact that humans can recognize the first time hearing or seeing something is a testament to the fact that we are constantly comparing what we perceive with previous experiences to make sense of the world around us.

At a high level, memory allows for three crucial processes: (1)~acquisition of new information (encoding), (2)~information retention over time (storage), and (3)~retrieval.
Through these processes, we can make sense of our present and take informed actions based on past observations.

Our learning framework explores such characteristics of memory.
Given a pair of views $x^1$ and $x^2$, while most SSL methods compare views directly or against learnable prototypes, we seek to design a task that forces the neural network to utilize previously experienced concepts as discriminative cues to learn representations that are invariant to view changes.
This task~\eqref{eq:cross-entropy} must produce consistent predictions for different views of an image $x$ based on the similarity perspectives of representation vectors stored in memory.
In other words, a pair of views $x^1$ and $x^2$ must have consistent similarity relationships to previous concepts experienced by the model.

In practice, the memory is a non-parametric distribution that stores encoded representations $z^{(\cdot)}$ from the current batch of image views.
To update the memory, we implement a FIFO protocol (First-In, First-Out), \ie, representations enter from one end of the memory and are discarded from the other.
This storage pattern preserves an \textit{ordering} bias in which one end of the memory holds recently updated representations while the other holds older ones.
Since representations constantly evolve during training, this ordering bias could drive the learning algorithm to give more weight to the recently remembered representations stored in one end of the memory.
We break the ordering dependency by introducing a stochastic component when retrieving representations from memory. We show that such a strategy regularizes the model and improves representations, \cf \Cref{sec:strategies-sample-mem-blocks}.

\subsection{Optimizing over Random Memory Blocks}
\label{sec:random-memory-blocks}

Inspired by \citepos{Silva2023} work on the random partition pretext task, we empirically found that applying a similar principle to our proposed memory component to break it into multiple disjoint subsets further improves performance and training stability, \cf \Cref{sec:strategies-sample-mem-blocks}.
Randomizing the memory representations into independent smaller blocks effectively mitigates the ordering bias that arises from inserting recent experiences into one end of the FIFO memory. This approach not only improves the overall performance but also enhances the training stability of the system.
Consequently, we transition from a single task over all representations in memory (ordered by insertion time) to a series of smaller tasks, each operating on a small subset of independent memory representations.

Let $\mathbb{M} = \left \{ M_1, M_2, \dots, M_{B} \right \}$ form a partition of the set $\mathcal{M}$, where $M_b$, for $b=1,2, \dots, B$ is a non-empty subset, a memory block containing randomly chosen representations sampled from $\mathcal{M}$.

The framework starts by computing the representation vectors $z^1 = f\left( x^1 \right )$ and $z^2 = g\left ( x^2 \right )$, independently, for each view $x^v$.
Then, we retrieve a memory block $M_b$ (a random subset from $\mathcal{M}$) and compute the similarity scores between the views and the memory block as
\begin{align}
\label{eq:view-memory-similarity-relationship}
p^1_s &= \operatorname{softmax}\left( \cos\left( z^1, M_b \right) / \tau_s \right), \\
p^2_t &= \operatorname{softmax}\left( \cos\left( z^2, M_b \right) / \tau_t \right),
\end{align}
\noindent where $\cos\left ( \cdot,\cdot \right )$ is the cosine similarity function, $\tau_s$ and $\tau_t$ are student and teacher temperature hyper-parameters, and $p^1_s$ and $p^2_t$ are the \textit{view-memory similarity relationship} obtained from comparing the views' representations $z^1$ and $z^2$ and the representations $m_k$ within a memory block $M_b$.

Intuitively, the term $\cos\left ( z^{(\cdot)}, M_b \right )$ compares what is being perceived at the moment, \ie, the current image views $x^{(\cdot)}$, with what has been experienced in the past, $M_b$, creating a view-memory similarity relationship $p^{(\cdot)}$.
Once we contrast the current views' and memory blocks' representations, we force the view-memory similarity relationship to be consistent using a regular cross-entropy loss, such as
\begin{equation}
    \label{eq:cross-entropy}
    \mathcal{L}_b \left(p_s^{1}, p_t^{2} \right) = - p_t^{2,b} \log \left ( p_s^{1,b}  \right ),
\end{equation}
\noindent where the subscript $b$ indexes the memory blocks $M_b$.
The overall loss is the aggregation over the memory block losses, \ie, $\mathcal{L} = \sum_b \mathcal{L}_b$.

Optimizing the loss function~\eqref{eq:cross-entropy} forces a consistent assignment of views from the perspective of the representations currently remembered by the model.
Given a memory block $M_b$ of size $N_b$, the problem can be seen as a $N_b$-way classification task where each representation in a memory block represents a different semantic perspective.
This way, to achieve consistency between the pair of views, the similarity relationship between what the network remembers and the different views of an image must be consistent.
Intuitively, if we consider $C$ as the number of hidden classes in $\mathcal{X}$ and that the memory $\mathcal{M}$ is large enough $K \gg C$ such that we can assume the memory holds a fair number of representations from each hidden category, we would strive for two main properties when optimizing the loss function~\eqref{eq:cross-entropy}: (1)~the view-memory similarity relationship of each view should be consistent and (2)~the view-memory similarities should be higher for cases where the current views and the recalled representations share semantic structure, \ie remembering from a previously experienced concept.

\section{Main Results}

We assess \methodname's representations on a broader set of computer vision benchmarks, focusing on the challenging scenario of transfer learning with frozen features.

\subsection{Transfer Learning}

We follow the transfer learning evaluation protocol proposed by \citet{Silva2023} based on $k$-NN.
We validate~\methodname's representations on \textit{eight} datasets across four different values of $k$ and report results on \cref{tab:knn-transfer}.
For a fixed value of $k=20$, \methodname's \textbf{achieve better transfer scores on five out of the eight (5/8) datasets, with $k$-NN performance gains of \textcolor{OliveGreen}{+2.6} and \textcolor{OliveGreen}{+4.6} on AirCraft and GTSRB datasets respectively}.
On average, over all datasets, \methodname outperforms competitors with \textbf{performance gains of nearly \textcolor{OliveGreen}{+2.5} for all values of $k$}. We report additional experiments on \Cref{tab:transferLearningFull} in \Cref{ap:extended-results-transfer-learning}.

\begingroup
\setlength{\tabcolsep}{2.5pt}
\begin{table*}[tb]
\caption[transfer learning with $k$-NN.]{\textbf{Transfer learning $k$-NN evaluation.} We report top-1 accuracy ($k=20$) for individual datasets and averages over all datasets for $k \in \{10,20,100, 200 \}$. Results for ViT-B/16.}
\label{tab:knn-transfer}
\centering
\sisetup{
    table-format=2.1,
    round-mode = places,
    round-precision=1,
    detect-all,
}
\small
\begin{sc}
\begin{tabular}{llSSSSSSSSSSSS}
\toprule
&& {Pets} & {Flowers} & {Aircraft} & {Cars} & {Country} & {Food} & {STL} & {GTSRB} & \multicolumn{4}{c}{Avg $@k$} \\
\cmidrule{3-10}\cmidrule{11-14}
{Methods} & {Epo.} & \multicolumn{8}{c}{results for $k=20$} & {10} & {20} & {100} & {200} \\
\midrule
MAE & 800 & 19.43308803 & 16.86453082 & 9.720972097 & 5.96940679 & 5.014218009 & 11.94455446 & 64.6125 & 27.56927949 & 20.87444177 & 20.14106871 & 16.88431841 & 15.18440999 \\
MoCo-v3 & 300 & 83.78304715 & 70.20653765 & 27.39273927 & 22.36040293 & 14.25118483 & 64.49108911 & 97.475 & 56.12826603 & 55.26877365 & 54.51103337 & 52.1680542 & 51.32303884 \\
DINO & 800 & 90.05178523 & 84.61538462 & 38.49384938 & 32.67006591 & \bfseries 15.91943128 & 70.71683168 & 98.8875 & 64.69517023 & 61.9921799 & 62.00625229 & 60.84304905 & 60.19904136\\
iBOT & 800 & 89.23412374 & 83.36314848 & 33.69336934 & 28.81482403 & 15.69668246 & \bfseries 72.58613861 & \bfseries 98.9875 & 63.0482977 & 60.823866 & 60.67801055 & 59.51137382 & 58.79899178\\
\rowcolor{highlight} Ours & 800 & \bfseries 91.60534206 & \bfseries 84.64791023 & \bfseries 41.13411341 & \bfseries 33.27944286 & 15.67772512 & 72.45148515 & 98.8 & \bfseries 69.31908155 & \bfseries 63.34501114 & \bfseries 63.36438755 & \bfseries 62.44759915 & \bfseries 61.83397873\\
\bottomrule
\end{tabular}
\end{sc}
\end{table*}
\endgroup

In addition to the non-parametric $k$-NN benchmark, we train logistic regression classifiers on top of the frozen features of the pre-trained ViT-B encoder.
In \Cref{tab:transfer-logreg}, we compare the performance of SSL methods on \textit{six} datasets. \methodname's representations \textbf{achieve higher transferable scores in four of the six datasets (4/6)}, highlighting the high transfer-learning power of \methodname's pre-trained representations.
\begingroup
\setlength{\tabcolsep}{4.5pt}
\begin{table}[tb]
\caption{\textbf{Transfer learning with logistic regression.} We report top-1 accuracy for logistic regression models trained on top of the frozen features of SSL ViTs pre-trained on the ImageNet-1M.}
\label{tab:transfer-logreg}
\centering
\sisetup{
    table-format=2.1,
    round-mode = places,
    round-precision=1,
    detect-all,
}
\small
\begin{sc}
\begin{tabular}{lSSSSSS}
\toprule
Method & {DTD} & {C100} & {GTSRB} & {Cars} & {Air} & {Pets}\\
\midrule
DINO & \bfseries 73.29787234 & 82.56 & 86.79334917 & 70.33951001 & 66.09660966 & 93.54047424 \\
iBOT & 71.75531915 & \bfseries 84.04 & 85.7719715 & 70.52605397 & 64.47644764 & 93.75851731 \\
\rowcolor{highlight} Ours & 71.70212766 & 83.08 & \bfseries 89.12114014 & \bfseries 73.73461012 & \bfseries 67.23672367 & \bfseries 94.16734805 \\
\bottomrule
\end{tabular}
\end{sc}
\end{table}
\endgroup

\subsection{Linear Evaluation}
\label{sec:linear-evaluation}

In~\Cref{tab:linear-probing}, we report in-domain linear evaluations for ViT-S/B backbones by training linear models with SGD, \cf \Cref{ap:linear-probing} for details on the protocol.
Additionally, we report $k$-NN performance on the full ImageNet-1M. \methodname performs on par with iBOT on both metrics, with a slight performance gain on ViT-B. We report the supervised baseline from \citepos{touvron2021training} work for reference.
\begingroup
\setlength{\tabcolsep}{2.5pt}
\begin{table}[tb]
\caption{\textbf{Linear probing and $k$-NN evaluation on ImageNet-1M.}}
\label{tab:linear-probing}
\centering
\sisetup{
    table-format=2.1,
    round-mode = places,
    round-precision=1,
    detect-all,
}
\small
\begin{sc}
\begin{tabular}{lllSS}
\toprule
Method & {Arch} & {Epo.} & {Linear} & {$k$-NN} \\
\midrule
MoCo-v3 & ViT-S/16 & 300 & 73.4 & {--} \\
DINO & ViT-S/16 & 300 & 76.2 & 72.8 \\
iBOT & ViT-S/16 & 300 & 77.4 & 74.6 \\
\rowcolor{highlight} Ours & ViT-S/16 & 300 & \bfseries 77.5 & \bfseries 74.7 \\
\midrule
DeiT (Sup) & ViT-S/16 & 800 & 79.8 & {--} \\
DINO & ViT-S/16 & 800 & 77.0 & 74.5 \\
iBOT & ViT-S/16 & 800 & \bfseries 77.9 & \bfseries 75.2 \\
\rowcolor{highlight} Ours & ViT-S/16 & 800 & 77.8 & 75.1 \\
\midrule
DeiT (Sup) & ViT-B/16 & 400 & 81.8  & {--} \\
MoCo-v3 & ViT-B/16 & 400 & 76.7 & {--} \\
NNCLR & ViT-B/16 & 1000 & 76.5 & {--} \\
DINO & ViT-B/16 & 400 & 78.2 & 76.1 \\
iBOT & ViT-B/16 & 400 & 79.5 & 77.1 \\
\rowcolor{highlight} Ours & ViT-B/16 & 400 & \bfseries 79.6 & \bfseries 77.2 \\
\bottomrule
\end{tabular}
\end{sc}
\end{table}
\endgroup

\subsection{Image Retrieval Benchmark}
Following previous work \citep{caron2021emerging,zhou2021ibot}, we evaluate \methodname's pre-trained representations on retrieval tasks based on \textit{landmark} and \textit{copy detection}.

\textbf{Image Retrieval.}
We consider the widely used revisited Oxford-5k and Paris-6k image retrieval datasets \citep{radenovic2018revisiting}, containing 3 distinct sets of increasing difficulty, each with query/database pairs.
In~\Cref{tab:image-retrieval}, we report mAP (mean average precision) on the Medium (M) and Hard (H) sets, ensuring fair comparisons with previous work.
For reference, we report the performance of a supervised method \citep{revaud2019learning} tailored for image retrieval tasks.

Among SSL methods, DINO is a strong baseline and beats iBOT in most instances.
\methodname's ViT-S surpasses DINO in all scenarios while our ViT-B pre-trained encoder outperforms DINO in \textbf{three out of the four (3/4) scenarios}, only losing in the Hard set of the Oxford-5k dataset by \textcolor{red}{-0.2}.

\begingroup
\setlength{\tabcolsep}{3.5pt}
\begin{table}[t]
\caption{\textbf{Image retrieval.} We report mAP on the revisited Oxford-5k and Paris-6k retrieval datasets for different SSL methods using pre-trained frozen features from different ViT backbones.}
\label{tab:image-retrieval}
\centering
\sisetup{
    table-format=2.1,
    round-mode = places,
    round-precision=1,
    detect-all,
}
\small
\begin{sc}
\begin{tabular}{l@{ }l@{ }l@{ }SSSSS}
\toprule
& & & \multicolumn{2}{c}{$\mathcal{RO}\textup{x}$} & \multicolumn{2}{c}{$\mathcal{R}\textup{Par}$} \\
\cmidrule{4-7}
Method & Arch & Epo. & {M} & {H} & {M} & {H} \\
\midrule
Sup & RN101+R-MAC & 100 & 49.8 & 18.5 & 74.0 & 52.1 \\
\midrule
MoCo-v3 & ViT-S/16 & 300 & 21.74 & 5.09 & 38.91 & 13.09 \\
DINO & ViT-S/16 & 800 & 37.2 & 13.9 & 63.1 & 34.4 \\
iBOT & ViT-S/16 & 800 & 36.61 & 12.96 & 61.49 & 34.06 \\
\rowcolor{highlight} Ours & ViT-S/16 & 800 & \bfseries 38.53 & \bfseries 15.86 & \bfseries 63.42 & \bfseries 34.78 \\
\midrule
MoCo-v3 & ViT-B/16 & 300 & 30.53 & 8.62 & 54.28 & 23.49 \\
DINO & ViT-B/16 & 400 & 37.38 & 13.73 & 63.5 & 35.6 \\
iBOT & ViT-B/16 & 400 & 36.84 & \bfseries 14.25 & 64.1 & 36.64 \\
\rowcolor{highlight} Ours & ViT-B/16 & 400 & \bfseries 39.32 & 14.12 & \bfseries 65.77 & \bfseries 38.09 \\
\bottomrule
\end{tabular}
\end{sc}
\end{table}
\endgroup

\textbf{Copy Detection.}
In addition, we consider the INRIA Copydays dataset \citep{douze2009evaluation} for evaluation on the copy detection task.
Following \citepos{zhou2021ibot} protocol, we report mAP on the ``strong'' subset without additional distractors in \Cref{tab:copy-detection}.
For ViT-B, \methodname increases upon the baseline performance from DINO and iBOT by \textcolor{OliveGreen}{\textbf{+0.8\%}} and \textcolor{OliveGreen}{\textbf{3.4\%}}, respectively.

\begingroup
\begin{table}[t]
\caption{\textbf{Copy detection.} We report mAP on the ``strong'' subset of the INRIA Copydays using frozen features from pre-trained ViTs.}
\label{tab:copy-detection}
\centering
\small
\begin{sc}
\begin{tabular}{lllS}
\toprule
{Method} & {Arch} & {Epo.}  & {mAP} \\
\midrule
DINO & VIT-S/16 & 800 & \bfseries 85.7 \\
iBOT & VIT-S/16 & 800 & 83.7 \\
\rowcolor{highlight} Ours & VIT-S/16 & 800 & 85.5 \\
\midrule
DINO & VIT-B/16 & 400 & 86.8 \\
iBOT & VIT-B/16 & 400 & 84.2 \\
\rowcolor{highlight} Ours & VIT-B/16 & 400 & \bfseries 87.6 \\
\bottomrule
\end{tabular}
\end{sc}
\end{table}
\endgroup

\subsection{Low-Shot and Long-Tailed Evaluation}

In \Cref{tab:low-shot-imnet}, we assess pre-trained representations on learning from a few labeled examples, where we consider $1\%$ and $10\%$ of the ImageNet labels.
Moreover, we measure the impact of using different evaluation protocols on low-shot classification by employing a non-parametric $k$-NN, a logistic regression estimator, and a linear model (single layer MLP) trained with SGD\@.
\textbf{\methodname outperforms DINO and iBOT in most setups}.
Interestingly, for ViT-S, training a linear model with SGD tends to underperform compared to logistic regression or even $k$-NN\@.
However, when more data or a more complex encoder is used, $k$-NN acts as a lower bound, while logistic regression and MLP alternate as the most effective evaluators.
\begingroup
\begin{table}[t]
\caption{\textbf{Low-shot classification on ImageNet-1M.} Evaluations on three protocols ($k$-NN, 1-layer MLP, and logistic regression) and two data regimes (1\% and 10\% of ImageNet-1M labels).}
\label{tab:low-shot-imnet}
\centering
\sisetup{
    round-mode = places,
    round-precision=1,
    detect-all,
}
\small
\begin{sc}
\begin{tabular}{lllSSS}
\toprule
Method & Arch & Protocol & {1\%}  & {10\%} \\
\midrule
DINO & VIT-S/16 & $k$-NN & 61.3 & 69.1 \\
iBOT & VIT-S/16 & $k$-NN & 62.486 & 70.112 \\
\rowcolor{highlight} Ours & VIT-S/16 & $k$-NN & \bfseries 62.582 & \bfseries 70.402 \\
DINO & VIT-S/16 & Linear & 60.5 & 71.0 \\
iBOT & VIT-S/16 & Linear & \bfseries 61.5 & 72.6 \\
\rowcolor{highlight} Ours & VIT-S/16 & Linear & 61.4 & \bfseries 72.6 \\
DINO & VIT-S/16  & LogReg & 64.5 & 72.2 \\
iBOT & VIT-S/16  & LogReg & 65.9 & \bfseries 73.4 \\
\rowcolor{highlight} Ours & VIT-S/16 & LogReg & \bfseries 65.898 & 73.154 \\
\midrule
DINO & VIT-B/16 & $k$-NN & 62.486 & 70.112 \\
iBOT & VIT-B/16 & $k$-NN & 66.298 & 72.892 \\
\rowcolor{highlight} Ours & VIT-B/16 & $k$-NN & \bfseries 68.804 & \bfseries 74.108 \\
DINO & VIT-B/16 & Linear & 66.2 & 74.244 \\
iBOT & VIT-B/16 & Linear & 68.2 & 75.698 \\
\rowcolor{highlight} Ours & VIT-B/16 & Linear & \bfseries 70.4 & \bfseries 76.4 \\
DINO & VIT-B/16 & LogReg & 67.082 & 74.18 \\
iBOT & VIT-B/16 & LogReg & 69.608 & 75.86 \\
\rowcolor{highlight} Ours & VIT-B/16 & LogReg & \bfseries 71.298 & \bfseries 76.33 \\
\bottomrule
\end{tabular}
\end{sc}
\end{table}
\endgroup

In \Cref{tab:low-shot-and-long-tail}, we consider long-tailed learning and challenging low-shot scenarios. We train linear models using frozen features on the ImageNet-LT dataset~\citep{liu2019large}, which is a highly unbalanced version of the ImageNet-1M\@.
We report top-1 accuracy on the ImageNet-LT \textit{balanced} test set. In addition, we report top-1 accuracy on balanced subsets of ImageNet-1M containing \textit{one}, \textit{two}, and \textit{four} randomly chosen examples per class. \methodname shows \textbf{significant learning efficiency on extreme low-shot scenarios and robustness to highly unbalanced data}.
\Cf \Cref{ap:low-shot-and-long-tail} for more details.

\begingroup
\setlength{\tabcolsep}{4.4pt}
\begin{table}[t]
\caption{\textbf{Low-shot and long-tailed evaluations.} We report Top-1 accuracy for ViT-B/16 on low-shot and long-tailed ImageNet.}
\label{tab:low-shot-and-long-tail}
\centering
\footnotesize
\begin{sc}
\sisetup{
    table-format=2.1,
    round-mode = places,
    round-precision=1,
    detect-all,
}
\begin{tabular}{lSSS{}S}
\toprule
 & \multicolumn{3}{c}{\# images per class} & {ImNet-LT}\\
\cmidrule{2-5}
& {1} & {2} & {4} & {top-1} \\
\midrule
MoCo-v3 & {37.7$\pm$ 0.3} & {47.8$\pm$ 0.6} & {54.8$\pm$ 0.2} & 56.68 \\
DINO & {39.2$\pm$ 0.4} & {49.3$\pm$ 0.8} & {57.6$\pm$ 0.4} & 63.676 \\
iBOT & {42.2$\pm$ 0.7} & {52.8$\pm$ 0.3} & {60.6$\pm$ 0.3} & 66.15 \\
\rowcolor{highlight} Ours & \bfseries {44.8$\pm$ 0.4} & \bfseries {56.3$\pm$ 0.3} & \bfseries {63.8$\pm$ 0.2} & \bfseries 67.898 \\
\bottomrule
\end{tabular}
\end{sc}
\end{table}
\endgroup

\subsection{Robustness Evaluation}

Vision models rely on foreground and background information when classifying objects in images.
Even when the correct object is present in an image, changes in the background may cause the network to classify the object incorrectly.
To understand how background-robustness in SSL methods, we follow the protocol from \citet{zhou2021ibot} and assess the robustness of pre-trained SSL representations against background changes using the ImageNet-9 (IN-9) dataset \citep{xiao2020noise}.
The IN-9 evaluation protocol masks/superimposes foregrounds on adversarially chosen backgrounds to define the following seven variants: Only-FG (OF), Mixed-Same (MS), Mixed-Rand (MR), Mixed-Next (MN), No-FG (NF), Only-BG-B (OBB), and Only-BG-T (OBT).
In the first four variants, the original foreground is kept while the background is modified. In the last three variants, the original foreground is masked.

In \Cref{tab:robustness-eval}, we report results for ViT-B backbones trained for \num{400} epochs and then evaluated using a linear head for \num{100} epochs.
Even though \methodname only trains on the \texttt{[CLS]} token of the ViT, it still surpasses iBOT, which performs MIM (Masked Image Modeling) on the patch tokens, \textbf{on four out of the seven (4/7) variants}.
\begingroup
\setlength{\tabcolsep}{3.5pt}
\begin{table}[t]
\caption{\textbf{Robustness evaluation against background changes.} ViT-B results on the IM-9 dataset over 7 variants of foreground/background mixing and masking.}
\label{tab:robustness-eval}
\centering
\small
\begin{sc}
\sisetup{
    table-format=2.1,
    round-mode = places,
    round-precision=1,
    detect-all,
}
\begin{tabular}{lSSSSSSSS}
\toprule
 & \multicolumn{7}{c}{Background changes} & {Clean}\\
\cmidrule{2-9}
 & {OF} & {MS} & {MR} & {MN} & {NF} & {OBB} & {OBT} & {IN-9} \\
\midrule
iBOT & 91.85 & 89.73 & 81.93 & 79.73 & 54.67 & 17.56 & 20.35 &  96.77\\
\rowcolor{highlight} Ours & 91.04 & \bfseries 90.15 & \bfseries 82.99 & \bfseries 80.40 & 53.38 & 15.80 & \bfseries 23.65 & \bfseries 97.60 \\
\bottomrule
\end{tabular}
\end{sc}
\end{table}
\endgroup

\subsection{Clustering Evaluation}

In addition to the supervised evaluations in~\Cref{sec:linear-evaluation}, we assess pre-trained representations using unsupervised metrics on the ImageNet-1\% and CIFAR-10 datasets in \Cref{tab:clutering-eval}.
\methodname \textbf{outperforms iBOT in all cases and performs comparably to DINO}.

\begingroup
\setlength{\tabcolsep}{3.2pt}
\begin{table}[t]
\caption{\textbf{Clustering evaluation}. We report (NMI) normalized mutual information, (AMI) adjusted mutual information, and (ARI) adjusted rand index.}
\label{tab:clutering-eval}
\centering
\small
\begin{sc}
\sisetup{
    table-format=2.1,
    round-mode = places,
    round-precision=1,
    detect-all,
}
\begin{tabular}{llSSS{ }SSS{ }SSS{ }}
\toprule
& & \multicolumn{3}{c}{ImageNet-1\%} & \multicolumn{3}{c}{CIFAR-10} \\
\cmidrule{3-8}
Method & {Arch} & {NMI} & {AMI} & {ARI} & {NMI} & {AMI} & {ARI} \\
\midrule
DINO & RN-50 & 69.2 & 46.2 & 21.7 & 39.6 & 39.5 & 28.0 \\
CARP & RN-50 & 70.3 & 48.0 & 23.9 & 49.0 & 48.9 & 38.7 \\
\midrule
DINO & ViT-B/16 & 79.07779248 & 64.3455682 & 38.1439685 & 58.74788555 & \bfseries 58.46283426 & \bfseries 27.43476552 \\
iBOT & ViT-B/16 & 81.3058528 & 68.05351608 & 42.04323482 & 57.74346597 & 57.45076803 & 26.78082051\\
\rowcolor{highlight} Ours & ViT-B/16 & \bfseries 81.72466395 & \bfseries 68.69777741 & \bfseries 44.06934797 & \bfseries 58.72199532 & 58.43683103 & 27.18218622\\
\bottomrule
\end{tabular}
\end{sc}
\end{table}
\endgroup

\subsection{Training Time and GPU Memory}

One important advantage of \methodname over other SSL methods based on ViTs is the trade-off between training resources (plus time) and performance.
DINO and iBOT learn prototypes using gradient descent. DINO trains \num{65536} prototypes, which translates into \num{16777216} extra trainable parameters, given the standard representation vector dimension of \num{256}.
On the other hand, \methodname avoids learning prototypes from scratch and implements a stochastic non-parametric memory component using representations from previous iterations, which require negligible extra computing memory since it does not receive gradient updates.
As shown in \Cref{tab:time-and-memory}, training \methodname for \num{800} epochs using a ViT-S backbone is nearly 9\% faster, requires 25.6\% less memory, achieves comparable linear probing on ImageNet-1M and better transfer performance on many datasets.

iBOT trains two sets of prototypes, each containing \num{8192} output neurons, and requires \num{4194304} trainable parameters to learn the prototype layers.
iBOT also uses the full output of the transformer, \ie, the \texttt{[CLS]} plus \textit{patch tokens}, which drastically increases its memory footprint and training time.
Differently, \methodname only trains on the \texttt{[CLS]} token of the ViT, still delivering good transferable performance in less time and with less memory.
Overall, \methodname \textbf{achieves the best trade-off between performance and training resources}.
All methods were trained on two 8-GPU V100 machines with a batch size of
\num{1024}.

\begingroup
\setlength{\tabcolsep}{2.5pt}
\begin{table}[tb]
\caption{\textbf{Training time and memory.} We compare performance ($k$-NN on ImageNet-1M), training time (hours), and memory requirements (Gigabytes) for SSL methods based on ViT-S/16 backbones pre-trained with a global batch size of \num{1024} images.}
\label{tab:time-and-memory}
\centering
\footnotesize
\begin{sc}
\begin{tabular}{llllllll}
\toprule
       & \multicolumn{2}{l}{100 epochs} & \multicolumn{2}{l}{300 epochs} & \multicolumn{2}{l}{800 epochs} &     \\
\cmidrule{2-7}
 & $k$-NN          & Time          & $k$-NN          & Time          & $k$-NN          & Time          & Mem \\
\midrule
DINO & 69.7 & 24.2h & 72.8 & 72.6h & 74.5 & 180.0h & 15.4G \\
iBOT & 71.5 & 24.3h & 74.6 & 73.3h & 75.2 & 193.4h & 19.5G \\
\rowcolor{highlight} Ours & 72.7 & 24.2h & 74.7 & 72.4h & 75.1 & 177.3h & 15.1G \\
\bottomrule
\end{tabular}
\end{sc}
\end{table}
\endgroup

\begin{figure}[tb]
\begin{center}
\centerline{\includegraphics[width=\columnwidth]{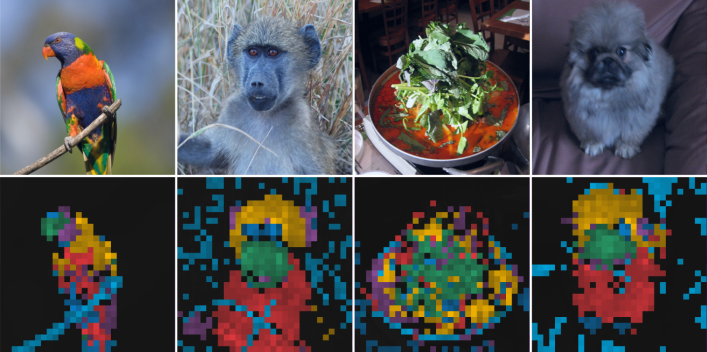}}
\caption{Visualization of \methodname's self-attention maps. Multiple heads are displayed in different colors.}
\label{fig:vis-sa-maps}
\end{center}
\vskip -0.4in
\end{figure}

\begin{figure}[tb]
\begin{center}
\centerline{\includegraphics[width=\columnwidth]{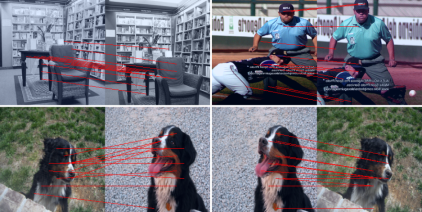}}
\caption{Sparse correspondence results for \methodname.}
\label{fig:sparse-corresp}
\end{center}
\vskip -0.2in
\end{figure}

\subsection{Visualizing Self-Attention Maps}

To analyze the internal representations of \methodname and to understand its powerful retrieval properties, we follow the protocols of \citet{caron2021emerging, zhou2021ibot} and visualize the attention maps of \methodname's pre-trained ViT-S encoder.
The \texttt{[CLS]} token is used as the query vector, and the visualizations are from different heads of the last layer displayed in various colors.

In \Cref{fig:vis-sa-maps}, we see the high attentive capabilities of \methodname.
For instance, in the first column, we see individual heads paying attention to different portions of the image, such as the bird's beak, different parts of its body, and the wood.
In the second column, we can visually isolate the monkey's top head, body, and face.
In the third column, multiple heads attend to different parts of the food. We present a detailed visual overview of \methodname's self-attention layers in \Cref{ap:vis-self-attn-maps}.

\subsection{Sparse Correspondence}

In \Cref{fig:sparse-corresp}, we evaluate \methodname's performance on the sparse correspondence task proposed by \citet{zhou2021ibot}.
The task is to match patch representations from two images.
In the first row of \Cref{fig:sparse-corresp}, we match patches from two views of the same image, while in the second row, we match patches from two distinct images of the same class.
Interestingly, even though \methodname does not train at patch-level representations, as iBOT, it still performs surprisingly well on dense prediction tasks such as sparse correspondence. We provide additional visualizations in \Cref{ap:space-correspondence}.

\section{Ablations}

In this section, we investigate the main components of \methodname. Unless otherwise specified, ablations experiments are pre-trained for \num{100} epochs using the ViT-S architecture varying hyperparameters according to the experiments.

\subsection{The Effect of the Memory Size}
\label{sec:ablations-memory-size}
One natural aspect of the proposed memory component in \Cref{sec:memory} is how its size affects the learned representation.
Intuitively, a large memory retains information for longer, allowing the model to compare the current image views to a broader distribution of remembered concepts.
Also, for fixed block sizes $N_b$, a large memory $M$ allows for more blocks, increasing signal processing during training. On the other hand, a smaller memory reduces the span and distribution of stored concepts.

In \Cref{tab:effects-of-memory-size}, we investigate the effect of the memory size on \methodname' performance. We fix the memory block size as $N_b=4096$ and vary the memory size $K$. We report top-1 accuracy using $k$-NN.
Experiments suggest that a larger memory benefits the learned representations up to a certain point where performance saturates. Based on these experiments, we set the memory size $K=65536$ unless otherwise stated.

\begingroup
\begin{table}[tb]
\caption{A larger memory benefits the learned representations.}
\label{tab:effects-of-memory-size}
\centering
\sisetup{
    table-format=2.1,
    round-mode = places,
    round-precision=1,
    detect-all,
}
\small
\begin{sc}
\begin{tabular}{lSSSSS}
\toprule
$K$ & {8192} & {16384} & {32768} & {65536} & {131072} \\
\midrule
$k$-NN & 67.824 & 69.872 & 70.529 & 71.862 & 71.924 \\
\bottomrule
\end{tabular}
\end{sc}
\end{table}
\endgroup

\subsection{The Effect of the Memory Block Size}

While the memory $M$ controls the span to which the model can remember previous concepts, the memory block size $N_b$ controls the dimensionality of the optimization problem. If $N_b$ is too small, we might limit the variability of concepts to which we compare the current image views to representations from past iterations. If $N_b$ is too large, we might encounter stability problems due to the weak self-supervised signal from the augmented views.

In \Cref{tab:effect-of-memory-block-size}, we investigate how the block size hyperparameter affects our framework. We report top-1 accuracy using $k$-NN for many configurations of $N_b$ while keeping the memory size $K=65536$.
Empirically, \methodname is robust to many configurations of $N_b$ and does not collapse even when using very large block sizes. The experiments suggest an optimal value of $N_b=16384$.

\begingroup
\setlength{\tabcolsep}{4.2pt}
\begin{table}[tb]
\caption{Larger block sizes $N_b$ benefit the learned representations.}
\label{tab:effect-of-memory-block-size}
\centering
\sisetup{
    table-format=2.1,
    round-mode = places,
    round-precision=1,
    detect-all,
}
\small
\begin{sc}
\begin{tabular}{lSSSSSSS}
\toprule
$N_b$ & {512} & {1024} & {2048} & {4096} & {8192} & {16384} & {32768} \\
\midrule
$k$-NN & 67.804 & 68.521 & 70.024 & 71.242 & 71.754 & 71.862  & 70.644\\
\bottomrule
\end{tabular}
\end{sc}
\end{table}
\endgroup

\subsection{Sampling Memory Blocks}
\label{sec:strategies-sample-mem-blocks}
In \Cref{tab:memory-block-sampling}, we compare different strategies to create memory blocks $M_b$ from the main memory $\mathcal{M}$. We consider two protocols. \textbf{Stochastic:} A Memory block contains randomly sampled (without replacement) representations from the memory. \textbf{Blockwise:} A memory block is a contiguous section of representations from the main memory.

Empirically, the Blockwise approach for memory blocks collapses regardless of the block size $N_b$.
This failure may be due to the FIFO update rule of the memory, which adds two properties to the learning mechanism.
First, FIFO updates add an ordering/sequence bias in the location of the representations in memory, in which representations from one end of the memory are older than representations on the other end.
Second, the FIFO updates shift (by a constant value) representations at each iteration towards the end of the memory. These update patterns make it easier for the network to \textit{overfit} to its memory and collapse the representations.

\begingroup
\begin{table}[tb]
\caption{Strategies for sampling memory blocks. We report $k$-NN top-1 accuracy for varying block sizes $N_b$.}
\label{tab:memory-block-sampling}
\centering
\sisetup{
    table-format=2.1,
    round-mode = places,
    round-precision=1,
    detect-all,
}
\small
\begin{sc}
\begin{tabular}{lSSSSS}
\toprule
Block Strategy & {512} & {1024} & {2048} & {4096} \\
\midrule
Stochastic & 67.8 & 68.5 & 70.0 & 71.2 \\
Blockwise & 0.1 & 0.1 & 0.1 & 0.1 \\
\bottomrule
\end{tabular}
\end{sc}
\end{table}
\endgroup

\section{Discussion}

\textbf{Connection with SSL Clustering Methods.}
Self-supervised clustering methods \citep{caron2020unsupervised,caron2021emerging,silva2022carl,silva2021consistent} usually learn a set of prototypes using gradient descent.
The biggest challenge in this setup is avoiding training collapse by solving the cluster assignment problem.
Alternative approaches \citep{PCL} use classic machine learning algorithms such as $k$-means to bootstrap centroids and pose classification problems over the views.
Regardless of the strategy, however, these methods usually require an explicit regularizer to avoid collapsed solutions.
We can view \methodname~from a clustering perspective where a set of randomly chosen embeddings are selected at each iteration to act as anchors or centroids.
Intuitively, \methodname's learning process may be seen as a form of randomly bootstrapping centroids from memory, which acts as a form of approximation of the training data embedding manifold.
Once initialized, these centroids are used to compute similarity scores across views.
This perspective hints that the memory size $K$ plays an important role and might depend on the number of hidden classes in the training dataset.
Intuitively, the memory must be large enough to hold a fair number of examples from each class, increasing the probability of a good initialization of the prototypes, \cf \Cref{tab:effects-of-memory-size}.

\section{Conclusion}

We presented \methodname, a memory-augmented self-supervised model for visual feature learning. \methodname draws on the intuitive properties of memory to use information from past training iterations to learn invariant representations for the current image views. \methodname offers interesting aspects such as (1)~the use of its memory component in the SSL task, (2)~the stochastic memory block sampling to regularize training, and (3)~the lack of additional regularizers to avoid collapse. Moreover, \methodname training architecture is simple and relatively cheaper to train. We provided many experimental results demonstrating our method's effectiveness in transfer and retrieval tasks.

\section*{Acknowledgements}

The computations were performed in part on resources provided by Sigma2---the National Infrastructure for High Performance Computing and Data Storage in Norway---through Project~NN8104K.
This work was funded in part by the Research Council of Norway, through its Centre for Research-based Innovation funding scheme (grant no.~309439), and Consortium Partners.
This study was financed in part by the Coordenação de Aperfeiçoamento de Pessoal de Nível Superior---Brasil (CAPES)---Finance Code 001.

\section*{Impact Statement}

``This paper presents work whose goal is to advance the field of Machine Learning. There are many potential societal consequences of our work, none which we feel must be specifically highlighted here.''\footnote{Verbatim statement according to the Call for Papers guidelines, \url{https://icml.cc/Conferences/2024/CallForPapers}.}

\bibliography{abrv,paper}
\bibliographystyle{icml2024}

\newpage
\appendix
\renewcommand\thetable{\Alph{section}.\arabic{table}}
\renewcommand\thefigure{\Alph{section}.\arabic{figure}}
\counterwithin{figure}{section}
\counterwithin{table}{section}

\onecolumn

\section{Evaluation Protocols}

\subsection{Transfer Learning with $k$-NN and logistic regression Models}
\label{ap:extended-results-transfer-learning}

\textbf{$k$-NN evaluation.}
We strictly follow the protocol and evaluation scripts from \citep{Silva2023} for transfer learning using $k$-NN classifiers.
We evaluate the following \num{8} datasets: Oxford-IIIT Pet, Oxford Flowers-102, AirCraft, Standard Cars, Country, Food-101, STL-10, and GTSRB.
For all experiments, we run $k$-NN with configurations of $k \in \{10,20,100, 200 \}$, and report the full results in~\Cref{tab:transferLearningFull} where we compare the performance of different SSL methods using the ViT-S and -B backbones for all datasets across \num{4} values of $k$.

\textbf{Logistic regression evaluation.} In \Cref{tab:transfer-logreg}, we report the transfer learning performance by training logistic regression models on top of the frozen features of the pre-trained ViT-B encoder. We use the \texttt{cyanure} library \citep{mairal2019cyanure} logistic regression implementation and the same set of hyper-parameters for all models. Below, we show the pseudo-code used to create the logistic regression classifier object using \texttt{cyanure}.

\begin{minted}[fontsize=\footnotesize]{python}
classifier = Classifier(loss="logistic", penalty="l2",
    solver="catalyst-miso", warm_start=False,
    max_iter=args.epochs,
    duality_gap_interval=10,
    fit_intercept=False,
    tol=1e-3,
    random_state=0,
    lambda_1=0.000002,
    lambda_2=0.000002)

classifier.fit(X, y)
\end{minted}

\begingroup
\setlength{\tabcolsep}{3.2pt}
\begin{table*}[tb]
\caption[Transfer learning evaluation.]{\textbf{Transfer learning evaluation}. We compare the top-1 $k$-NN accuracy of 9 SSL methods on 8 datasets. We report results for $k \in \{10,20,100, 200 \}$.}
\label{tab:transferLearningFull}
\centering
\sisetup{
    table-format=2.1,
    round-mode = places,
    round-precision=1,
    detect-all,
}
\small
\begin{sc}
\begin{tabular}{llSSSSSSSSSSSSSSSS}
\toprule
& & \multicolumn{4}{c}{Oxford-IIIT Pet} & \multicolumn{4}{c}{Oxford Flowers-102} & \multicolumn{4}{c}{Aircraft} & \multicolumn{4}{c}{Stanford Cars} \\
\cmidrule{3-18}
Method & {Arch} & {10} & {20} & {100} & {200} & {10} & {20} & {100} & {200} & {10} & {20} & {100} & {200} & {10} & {20} & {100} & {200} \\
\midrule
MoCo-v3 & ViT-S/16 & 34.5053148 & 34.58708095 & 33.30607795 & 30.85309349 & 51.26036754 & 50.18702228 & 41.64904863 & 37.1442511 & 16.89168917 & 17.52175218 & 16.08160816 & 15.42154215 & 9.476433279 & 9.376943166 & 8.954110185 & 8.245243129\\
DINO & ViT-S/16 & 91.79612974 & 91.19651131 & 90.65140365 & 90.24257291 & 83.39567409 & 82.51748252 & 81.60676533 & 82.02959831 & 39.9339934 & 40.0840084 & 35.94359436 & 33.39333933 & 27.84479542 & 27.85723169 & 27.24785474 & 26.47680637\\
iBOT & ViT-S/16 & 91.82338512 & 91.44180976 & 90.76042518 & 90.51512674 & 83.42819971 & 81.98080989 & 80.89120182 & 81.18393235 & 39.6939694 & 39.27392739 & 36.03360336 & 32.76327633 & 25.83012063 & 25.46946897 & 24.67354807 & 23.21850516\\
\rowcolor{highlight} Ours & ViT-S/16 & 91.52357591 & 91.8506405 & 90.43336059 & 90.46061597 & 84.46901935 & 83.57456497 & 82.64758497 & 83.08668076 & 38.61386139 & 37.92379238 & 35.19351935 & 32.52325233 & 30.48128342 & 31.04091531 & 29.85947022 & 29.05111305 \\
\midrule
MAE & ViT-B/16 & 20.95938948 & 19.43308803 & 15.69910057 & 14.30907604 & 18.71849081 & 16.86453082 & 10.18051716 & 9.20474874 & 9.480948095 & 9.720972097 & 8.310831083 & 6.600660066 & 6.33005845 & 5.96940679 & 4.825270489 & 4.763089168 \\
MoCo-v3 & ViT-B/16 & 83.42872717 & 83.78304715 & 81.43908422 & 80.40337967 & 74.92275167 & 70.20653765 & 64.30313872 & 65.99447065 & 28.47284728 & 27.39273927 & 23.10231023 & 21.57215722 & 23.36774033 & 22.36040293 & 21.34062927 & 20.0472578 \\
DINO & ViT-B/16 & 90.40610521 & 90.05178523 & 88.38920687 & 88.30744072 & 85.65620426 & 84.61538462 & 83.94860953 & 84.42023093 & 38.40384038 & 38.49384938 & 34.50345035 & 32.16321632 & 32.32185052 & 32.67006591 & 31.06578784 & 29.97139659 \\
iBOT & ViT-B/16 & 89.09784682 & 89.23412374 & 87.98037612 & 87.92586536 & 84.51780777 & 83.36314848 & 82.58253375 & 83.1842576 & 35.13351335 & 33.69336934 & 30.96309631 & 28.59285929 & 28.6158438 & 28.81482403 & 27.83235916 & 26.61360527 \\
\rowcolor{highlight} Ours & ViT-B/16 & 91.8506405 & 91.60534206 & 90.92395748 & 90.8967021 & 85.31468531 & 84.64791023 & 84.25760286 & 84.43649374 & 41.97419742 & 41.13411341 & 36.60366037 & 34.62346235 & 33.04315384 & 33.27944286 & 32.90635493 & 32.42134063 \\
\midrule
& & \multicolumn{4}{c}{Country-211} & \multicolumn{4}{c}{Food-101} & \multicolumn{4}{c}{STL-10} & \multicolumn{4}{c}{GTSRB} \\
\cmidrule{3-18}
Method & {ep} & {10} & {20} & {100} & {200} & {10} & {20} & {100} & {200} & {10} & {20} & {100} & {200} & {10} & {20} & {100} & {200} \\
\midrule
MoCo-v3 & ViT-S/16 & 7.563981043 & 7.521327014 & 7.165876777 & 6.592417062 & 28.57029703 & 31.12475248 & 34.29306931 & 34.06336634 & 66.75 & 67.1 & 64.3625 & 62.3 & 38.22644497 & 38.36104513 & 37.68804434 & 36.75376089\\
DINO & ViT-S/16 & 14.96208531 & 15.09478673 & 15.57345972 & 15.72985782 & 69.06930693 & 69.31485149 & 67.81386139 & 66.53861386 & 98.4375 & 98.425 & 98.225 & 98.175 & 60.59382423 & 61.14806017 & 61.39350752 & 60.59382423\\
iBOT & ViT-S/16 & 14.55924171 & 14.83412322 & 15.31753555 & 15.42180095 & 70.2019802 & 70.51881188 & 68.84356436 & 67.0970297 & 98.825 & 98.7875 & 98.55 & 98.475 & 61.52019002 & 61.95566112 & 61.63895487 & 60.71258907 \\
\rowcolor{highlight} Ours & ViT-S/16 & 14.53554502 & 14.97156398 & 15.45971564 & 15.56872038 & 69.73069307 & 70.28118812 & 68.9029703 & 67.65940594 & 98.275 & 98.3125 & 98.15 & 97.95 & 64.89311164 & 65.65320665 & 65.17814727 & 64.19635788 \\
\midrule
MAE & ViT-B/16 & 5.018957346 & 5.014218009 & 4.289099526 & 3.981042654 & 11.1049505 & 11.94455446 & 12.38415842 & 11.76633663 & 66.8 & 64.6125 & 54.825 & 48.2375 & 28.58273951 & 27.56927949 & 24.56057007 & 22.6128266\\
MoCo-v3 & ViT-B/16 & 14.2464455 & 14.25118483 & 13.13270142 & 12.51658768 & 64.22178218 & 64.49108911 & 62.19009901 & 60.28118812 & 97.6625 & 97.475 & 96.7375 & 95.6125 & 55.82739509 & 56.12826603 & 55.0989707 & 54.1567696\\
DINO & ViT-B/16 & 15.45971564 & 15.91943128 & 16.14218009 & 16.28909953 & 70.41584158 & 70.71683168 & 69.16831683 & 67.63168317 & 98.8875 & 98.8875 & 98.8 & 98.7 & 64.38638163 & 64.69517023 & 64.72684086 & 64.10926366\\
iBOT & ViT-B/16 & 15.36492891 & 15.69668246 & 16.08056872 & 16.14218009 & 72.0039604 & 72.58613861 & 70.41980198 & 68.8039604 & 98.975 & 98.9875 & 98.8625 & 98.8125 & 62.88202692 & 63.0482977 & 61.36975455 & 60.31670625 \\
\rowcolor{highlight} Ours & ViT-B/16 & 15.40758294 & 15.67772512 & 16.20379147 & 16.21800948 & 72.06336634 & 72.45148515 & 71.06138614 & 69.67524752 & 98.7375 & 98.8 & 98.6375 & 98.475 & 68.36896279 & 69.31908155 & 68.98653998 & 67.92557403 \\
\bottomrule
\end{tabular}
\end{sc}
\end{table*}
\endgroup

\subsection{Linear Probing and $k$-NN Evaluations on ImageNet}
\label{ap:linear-probing}

\textbf{Linear probing on ImageNet-1M.}
We closely follow the protocol and code scripts from \citep{zhou2021ibot} to train linear classifiers on the ImageNet-1M dataset on top of frozen features from pre-trained SSL methods. The evaluation script trains linear models with SGD, sweeps over different learning rates, and outputs the best model.

\subsection{Low-Shot and Long-Tailed Evaluations}
\label{ap:low-shot-and-long-tail}

\textbf{Low-shot on ImageNet.}

Due to reproducibility issues, we adapted the linear probing evaluation script and reran the low-shot classification experiments for DINO, iBOT, and MoCo-v3 using available subsets for 1\% and 10\% ImageNet labeled images from \citet{chen2020simple}, \cf \Cref{tab:low-shot-imnet}. Likewise, the evaluation script trains linear classifiers with SGD and sweeps over multiple learning rates. For low-shot evaluations using logistic regression on top of the frozen features, we use the \texttt{cyanure} library \citep{mairal2019cyanure}.

We train linear models with SGD on balanced subsets of the ImageNet dataset, where we allow a fixed number of examples per class. In \Cref{tab:low-shot-and-long-tail}, we report top-1 accuracy for three versions of the ImageNet data where only one, two, and four images are randomly sampled per class. We repeat the experiments 5 times and report top-1 accuracy and standard deviations.

\textbf{Long-tailed learning on ImageNet.}
To validate \methodname representations on highly unbalanced data, we train linear models (single layer MLPs) on the ImageNet-LT dataset~\citep{liu2019large}, which was designed as a long-tailed version of the ImageNet-2012. Its sampling strategy follows a Pareto distribution with a power value $\lambda=6$.
The ImageNet-LT contains \num{115.8}K images with a maximum of \num{1280} and a minimum of five images per class. In \Cref{tab:low-shot-and-long-tail}, we report performance (top-1 accuracy) on the balanced ImageNet-LT test set.

\section{Implementation Details}
\label{ap:implementation-details}

We train a joint-embedding teacher-student architecture using ViTs as backbones. We create multiple views of an image using different augmentation protocols. At each training iteration, we create 12 views from an image, 2 global views, each of size $224 \times 224$, and 10 local views, each of size $96 \times 96$. We follow the same augmentation protocol previously utilized by \citep{grill2020bootstrap}, namely a combination of color jittering,
Gaussian blur, solarization, and random crop.

The student $f$ and teacher $g$ branches have different ViT encoders and projection heads. The projection head follows the same architecture proposed by \citep{caron2020unsupervised},  \ie, a multilayer perceptron (MLP) with 3 layers, hidden size is 2048-d, and Gaussian error linear units (GELU) activations. Only the student branch receives gradient updates. The teacher branch is updated following an exponential moving average from the student's network weights.

We only consider the \texttt{[CLS]} token from the Transformer encoder. For reference, the ViT-B encodes image views to representation vectors of 768-d, which are then projected to a lower 256-d and normalized to have a unit hypersphere.

The memory $M$ is a non-differentiable container that holds representations at each training iteration and is updated following a FIFO (First-In, First-Out) strategy. The memory size is set to $K=65536$ following the ablations experiments in \Cref{sec:ablations-memory-size}. Before optimization, the view-memory similarity distribution is split into disjoint subsets called memory blocks, each of size $N_b=16384$, \Cf \Cref{sec:random-memory-blocks}.

\methodname is trained with the AdamW optimizer \citep{loshchilov2018fixing}, learning rate \num{1e-5}, and a global batch size of \num{1024}. The learning rate follows a cosine decay without warmup towards \num{1e-06}. Following \citep{caron2021emerging}, the weight decay follows a cosine schedule from \num{0.04} to \num{0.4}. The student temperature is set to $\tau_s = 0.1$, and the teacher temperature $\tau_t$ is warmed up from \num{0.04} to \num{0.07} in the first \num{30} epochs.

\subsection{PyTorch Style Pseudo-code}
\label{ap:pseudocode}

\begin{minted}[fontsize=\footnotesize]{python}
# D: Images' representation dimensionality
# K: Memory size
# NB: Memory block size
# B: Number of memory blocks
# N: Batch size
# z_i: Representation vector from the student encoder
# w_i: Representation vector from the teacher encoder

memory = torch.randn(D, K)
memory = F.normalize(memory, dim=0)

for x1, x2 in loader:
    # student and teacher branches
    z1, w1 = f(x1), g(x1) # [N, D]
    z2, w2 = f(x2), g(x2) # [N, D]

    p1, p2 = matmul(z1, memory), matmul(z2, memory) # [N, K]
    q1, q2 = matmul(w1, memory), matmul(w2, memory) # [N, K]

    # sample cluster indices with no replacement
    rand_proto_ids = torch.randperm(K)
    split_embed_ids = stack(split(rand_proto_ids, NB))

    ps, qs = [], []
    for p, q in zip([p1, p2], [q1, q2]):
        p_mb = fetch_mem_block(p, split_embed_ids)
        q_mb = fetch_mem_block(q, split_embed_ids)

        ps.append(p_mb)
        qs.append(q_mb)

    ps, qs = torch.cat(ps, dim=0), torch.cat(qs, dim=0)
    loss = loss_fn(ps, qs)

    # update memory
    enqueue(memory, w1)
    dequeue(memory)

    # gradient descent steps


def loss_fn(ps, qs):
    for i in range(len(ps)):
        for j in range(len(qs)):
            if i == j:
                continue
            consistency += cross_entropy(ps[i], qs[j])
            terms += 1
    consistency /= terms
    return consistency

def cross_entropy(p, q):
    p = torch.log_softmax(p, dim=-1)
    q = torch.softmax(q, dim=-1)

    loss = torch.sum(-q * p, dim=-1)
    return loss

def fetch_mem_block(logits, proto_ids):
    logits_gr = logits[:, proto_ids.flatten()]
    logits_gr = logits_gr.split(NB, dim=1)
    logits_gr = torch.cat(logits_gr, dim=0)
    return logits_gr # [N * B, NB]

\end{minted}

\section{Additional Results}
\subsection{Visualizing Self-Attention Maps}
\label{ap:vis-self-attn-maps}

\begin{figure}[tb]
\begin{center}
\centerline{\includegraphics[width=\columnwidth]{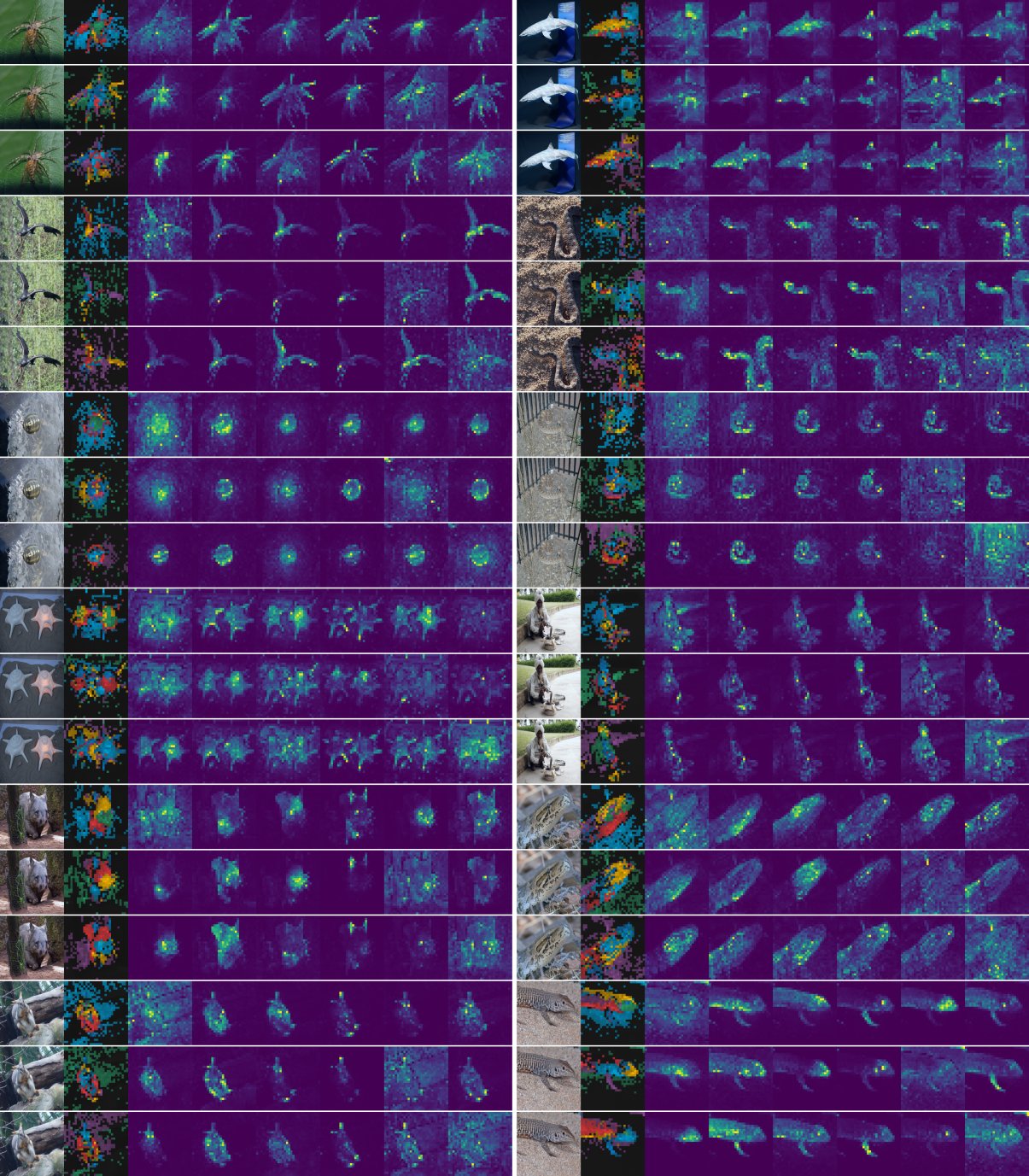}}
\caption{\textbf{Visualizing self-attention maps.}  From top to bottom, in each triplet of rows, we report qualitative evaluations for \methodname, iBOT, and DINO.  The columns show multiple attention heads of the last layer.}
\label{fig:comparing-attn-maps}
\end{center}
\end{figure}

We provide additional self-attention visualizations in~\Cref{fig:comparing-attn-maps}. We strictly follow the generating scripts from \citet{zhou2021ibot}, and display attention maps from pre-trained ViT-S backbones using images sampled from the validation set of the ImageNet-1M dataset, hence not used for training. In \Cref{fig:comparing-attn-maps}, for each image, we show attention maps from \methodname, iBOT, and DINO in this order from top to bottom. The protocol uses the \texttt{[CLS]} token as a query to extract attention maps over multiple heads of the last layer. \methodname learns comparable attentive maps to iBOT and DINO, where we can see the attention maps segmenting the object in the image and different heads paying attention to different features in the image.

\subsection{Sparse Correspondence}
\label{ap:space-correspondence}
\begin{figure}[tb]
\begin{center}
\centerline{\includegraphics[width=\columnwidth]{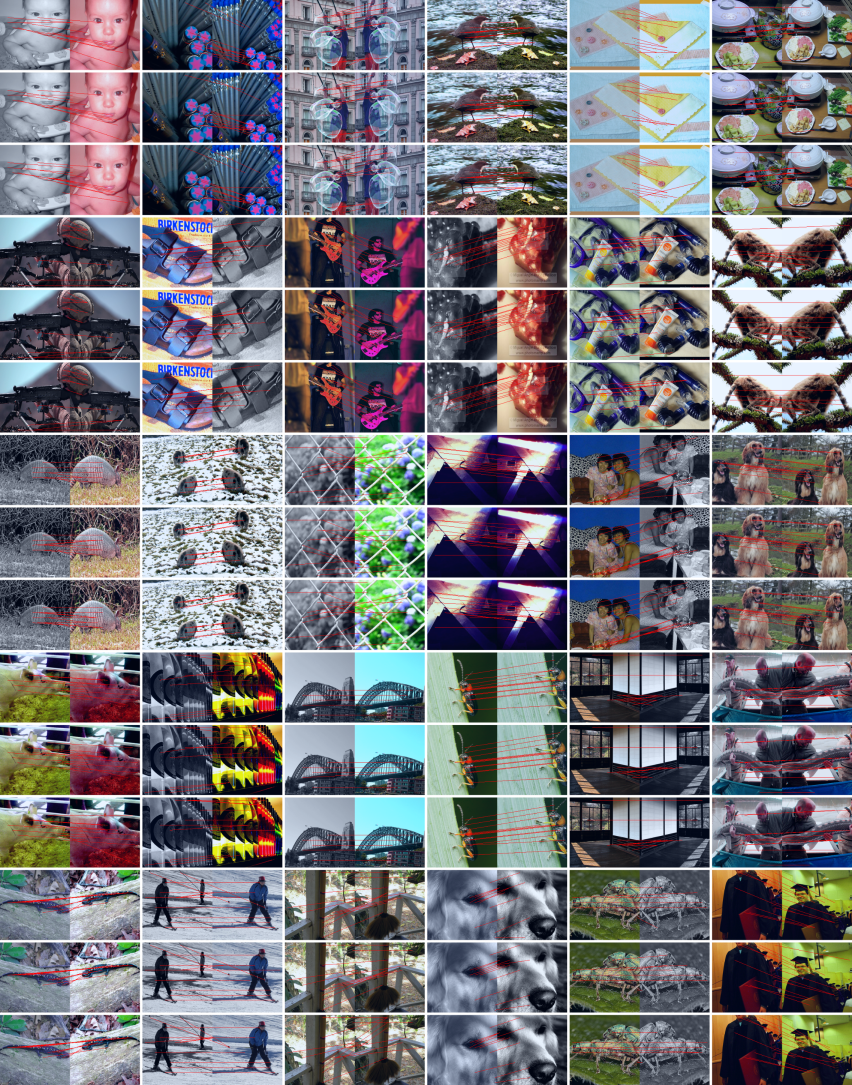}}
\caption{\textbf{Visualization for sparse correspondence.} We assess the ability to match local embeddings using pairs of views from the same image. From top to bottom, in each triplet of rows, we report qualitative evaluations for \methodname, iBOT, and DINO.
}
\label{fig:instance-level-correspondence}
\end{center}
\end{figure}

\begin{figure}[tb]
\begin{center}
\centerline{\includegraphics[width=\columnwidth]{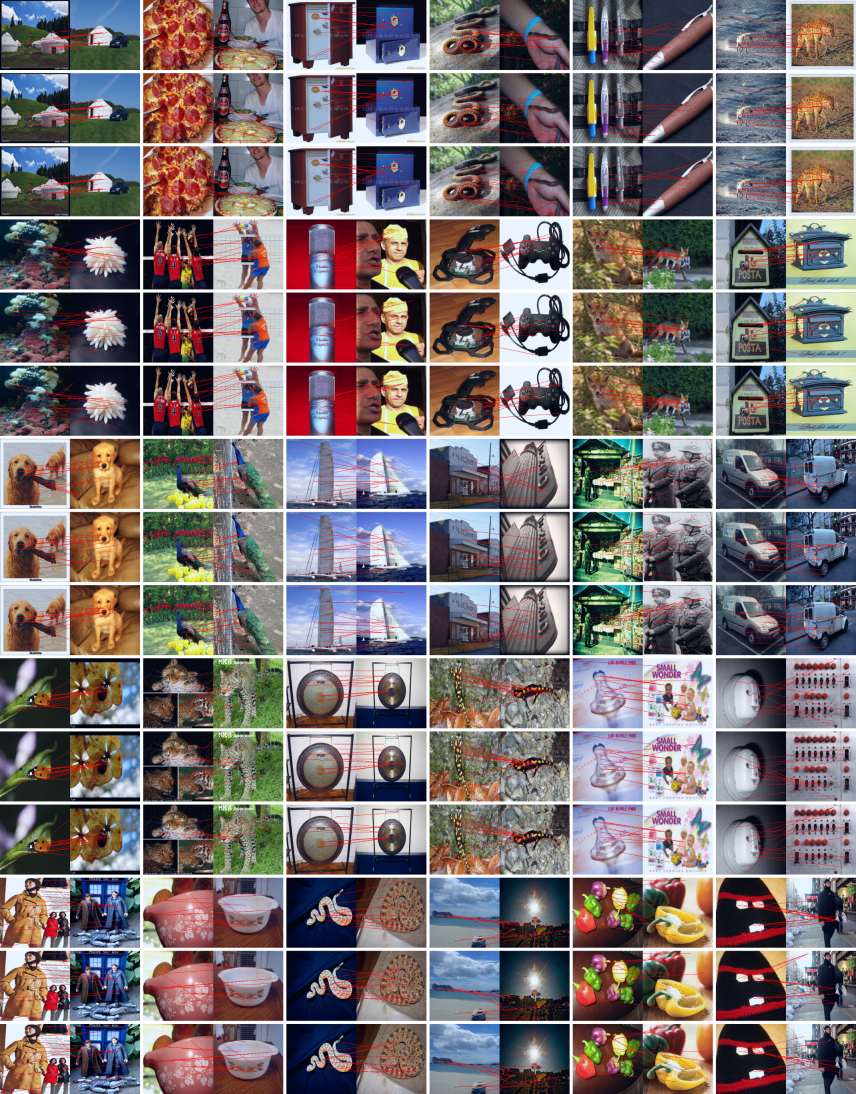}}
\caption{\textbf{Visualization for sparse correspondence.} We assess the ability to match local embeddings using a pair of images from the same class. From top to bottom, in each triplet of rows, we report qualitative evaluations for \methodname, iBOT, and DINO.
}
\label{fig:class-level-correspondence}
\end{center}
\end{figure}

We follow the sparse correspondence evaluation protocol proposed by~\cite {zhou2021ibot}, where the task is to match patch embeddings from the last layer of the ViT.
We qualitatively compare \methodname against iBOT and DINO using ViT-S/16 backbones pre-trained for \num{800}.
We consider two cases: (1)~the pair of matching images contain views extracted from the same image (\Cref{fig:instance-level-correspondence}) and (2)~the pair contains two distinct images from the same class (\Cref{fig:class-level-correspondence}).
The protocol matches local embeddings from two images, and at most $14 \times 14$ matched pairs can be extracted with a ViT-S. The evaluation script displays the 12 correspondences with the highest self-attention scores.
In~\Cref{fig:class-level-correspondence}, we show examples of feature correspondences for image pairs drawn from a wide variety of classes containing buildings, animals, humans, vehicles, and other objects. \methodname can extract mostly correct correspondences despite augmentations on scale and color.

\end{document}